\documentclass[conference]{IEEEtran}
\IEEEoverridecommandlockouts
\usepackage{cite}
\usepackage{amsmath,amssymb,amsfonts}
\usepackage{algorithmic}
\usepackage{graphicx}
\usepackage{textcomp}
\usepackage{xcolor}
\usepackage{subfigure}
\usepackage{soul}
\def\BibTeX{{\rm B\kern-.05em{\sc i\kern-.025em b}\kern-.08em
    T\kern-.1667em\lower.7ex\hbox{E}\kern-.125emX}}
\begin{document}

%\title{Augmenting Vital Signs Data from Healthy Individuals using Latent Modeling\\
%\title{Latent Modeling to Represent Clinical Conditions on Vital Signs from Healthy Individuals\\
\title{Representing Clinical Conditions on Vital Signs from Healthy Individuals using Latent Modeling\\
\thanks{This work was funded by the Engineering and Physical Sciences Research Council, grant number EP/X028631/1: ATRACT: A Trustworthy Robotic Autonomous system to support Casualty Triage.\\979-8-3315-1213-2/25/\$31.00 ©2025 IEEE}
}

\author{\IEEEauthorblockN{Rafael Pina}
\IEEEauthorblockA{\textit{Institute for Digital Technologies} \\
\textit{Loughborough University London}\\
London, United Kingdom \\
r.m.pina@lboro.ac.uk}
\and
\IEEEauthorblockN{Varuna De Silva}
\IEEEauthorblockA{\textit{Institute for Digital Technologies} \\
\textit{Loughborough University London}\\
London, United Kingdom \\
v.d.de-silva@lboro.ac.uk}
\and
\IEEEauthorblockN{Mindula Illeperuma}
\IEEEauthorblockA{\textit{Institute for Digital Technologies} \\
\textit{Loughborough University London}\\
London, United Kingdom \\
k.m.illeperuma@lboro.ac.uk}
}

\maketitle

\begin{abstract}
Machine learning can be crucial to help scale complex signal processing applications in scenarios such as healthcare. However, these machine learning models need rich datasets to be trained and there are often cases where it is not possible to access representative datasets. In this paper, we propose a deep generative model based on conditional variational autoencoders with the objective of augmenting the vital signs of healthy individuals in a way that mimics the patterns of a certain clinical condition. More specifically, we use a publicly available ICU (Intensive Care Unit) dataset to train our model and then evaluate it using the vital data that we have collected from healthy individuals. Our results demonstrate that the proposed model can not only learn the underlying dynamics of the ICU data but, more importantly, can reshape our collected data from healthy individuals in a way that is aligned with the vital signs of a certain clinical condition. We propose a distance metric that shows how our model can generate samples that are more aligned with the intended clinical labels when compared to the tested baselines. 
\end{abstract}


\section{Introduction}
Healthcare systems are one of the major beneficiaries of digital signal processing \cite{khan_2020_an,eeg_signal_processing_2023,biomedical_signal_proc_2023}. Signal processing techniques for vital sign monitoring and evaluation have been well studied area, and provide the foundation for many smart healthcare applications such as monitoring elderly, or heart conditions through the integration of machine learning techniques \cite{liu_intelligent_dsp_2022,amin_signal_proc_elderly_2016,mou_2024_eletronics,cuffless_bp_2021,cvae_2024_condition_monitor}. Monitoring vital signs such as heart and breathing rates or ECG, is a key technique used in Intensive Care Units (ICUs) to assist physicians, and surgeons in surgical theaters \cite{evans_vital_signs_2001}.   

There are also many critical healthcare applications in which vital sign monitoring and real-time processing and machine learning based predictions can be very useful. Consider the triaging of wounded citizens in a disaster zone, or wounded soldiers in a battlefield. Real-time signal analysis from wearable sensors can be extremely useful in saving lives in these situations \cite{carius2022battlefield,zhang_disaster_monitor_2023}. However, to utilise machine learning techniques to achieve predictive applications such as triaging in such settings is extremely difficult due to the difficulty of collecting representative data. 

Generative Artificial Intelligence (GenAI) has currently received significant attention from the research communities including the signal processing community \cite{wang_rf_genai_2025,huynh_dsp_genai_2024,wang_genai_wireless_sense_2024}. However, most of the GenAI applications have focused around text and image data, which are relatively easier to collect in massive amounts compared with vital sign signals pertaining to rare events. Sourcing large amounts of relevant data for healthcare applications is a massive challenge. On the other hand, data from healthcare applications is inherently multimodal, i.e., vital signs can come from various different modalities such as heart rate, breathing patterns, ECG, visual, radio frequency. In this paper, we work towards addressing this gap for a branch of signal processing applications of vital signs. 
\begin{figure}
    \centering
    \includegraphics[width=0.8\columnwidth]{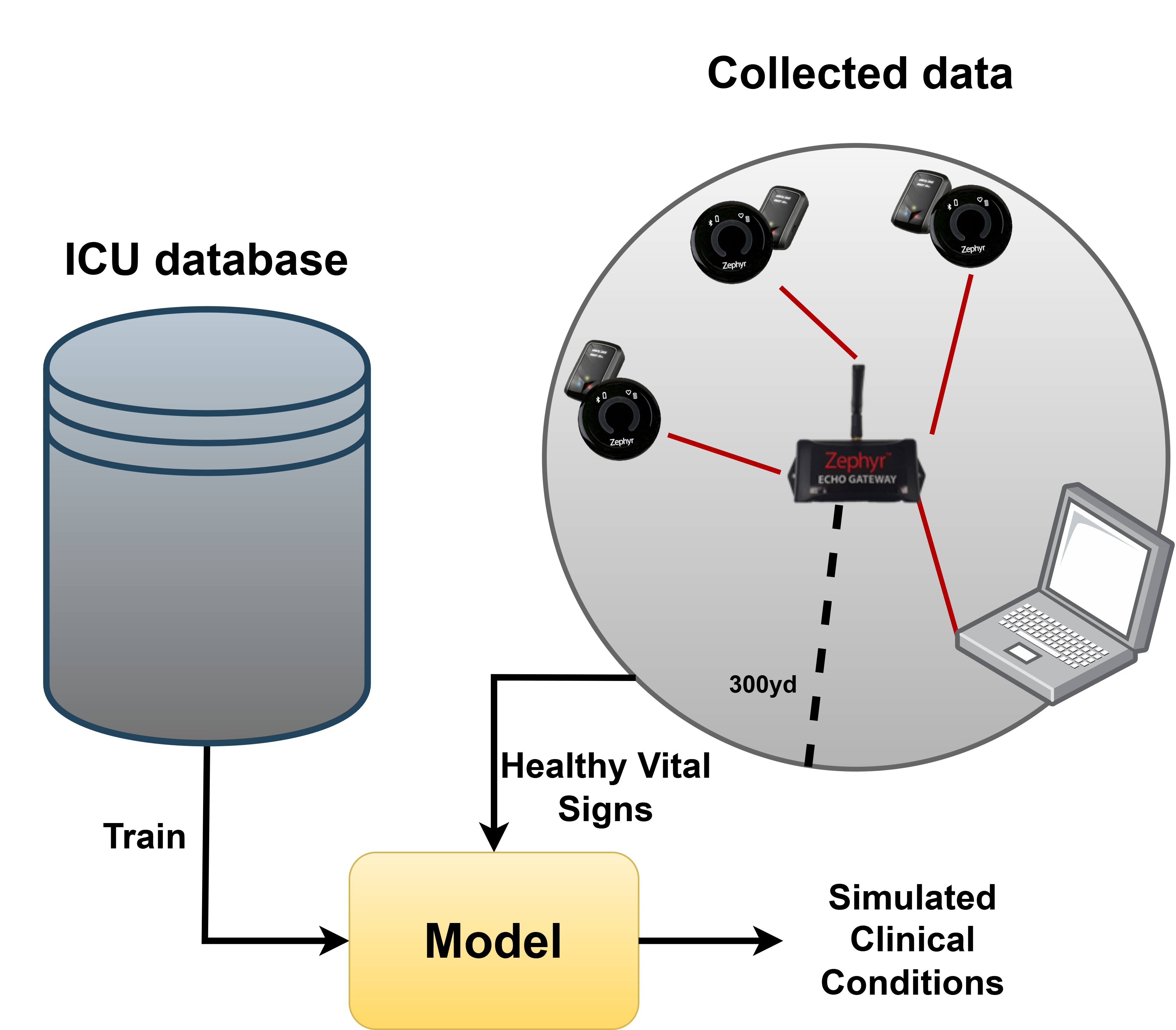}
    \caption{Diagram depicting our approach combining an ICU dataset and our collected data using Zephyr BioHarness modules \cite{zephyrtechnology_2012_bioharness} to simulate clinical conditions from healthy vital data.}
    \label{fig:psm_multi_diagram}
\end{figure}

Funded by the Engineering and Physical Sciences Research Council of UK, within the ATRACT project we consider the application of triaging wounded soldiers in a battlefield. Soldiers often wear multiple vital sign sensors. For triaging using machine learning, under different acute conditions we need to have data pertaining to that setting. However, collecting such data is prohibitive due to security, ethical and practical considerations. To solve that issue, we collect data from healthy individuals while having their movements, and then seek to augment the vital sign signals with the patterns from ICU patients. We propose a novel conditional variational autoencoder (CVAE) architecture to generate vital sign signals. The contributions of this paper are as follows:

\begin{itemize}
    \item We formulate a CVAE architecture that is capable of augmenting vital sign signals with acute medical conditions seen in ICU patients. The proposed model is conditioned on the required medical condition to simulate, and regularised for stable generation of signals. 
    \item We propose an objective method to evaluate the quality of signals that are generated from the deep learning architecture, for benchmarking purposes. 
\end{itemize}

The rest of this paper is organised as follows: we present the details of the data collection process, and the ICU dataset used for this work, in section \ref{sec:preliminaries}. The proposed variational architecture along with the evaluation method is presented in section \ref{sec:methods}, followed by experimental results and discussion in section \ref{sec:results}. We then conclude the paper in section \ref{sec:conclusion}.  

\section{Preliminaries}\label{sec:preliminaries}
\subsection{Data Collection}
We collected data from healthy individuals during normal physical activity that does not require intensive effort such as crawling and walking. To capture the vital signs of the participants, we used the Zephyr Technology BioHarness 3.0 BioModule \cite{zephyrtechnology_2012_bioharness} (Fig. \ref{fig:psm_multi_diagram}). This module can be attached to a strap placed around the chest of the participants, in direct contact with their skin allowing to record vital signals such as heart rate, breathing rate, acceleration, position and posture. 

We connected all the participants wearing the BioModules to a central laptop with the Zephyr OmniSense software running. With this software and the modules connected, it is possible to live-track the vital signs and performance of the participants. Although the modules only provide a short range, by using an ECHO gateway repeater we can extend the range of the signal significantly. The gateway used allows to establish a radio network following 2.4GHz 802.115.4, allowing a range of 300 yards that is covered by our radio network to which the BioModules can be connected. This scheme can be found in Fig. \ref{fig:psm_multi_diagram} (similar to a PSM configuration from \cite{zephyrtechnology_2014_psm}).

\subsection{The MIMIC Database}\label{sec:mimic-db}
To train the model proposed in this paper, we have selected a dataset from the MIMIC database \cite{moody_data_1996,goldberger_2000_physiobank}. We opted to use the MIMIC-I database since that is readily available out of the box (in the future, we intend to extend to the more recent versions). This dataset is formed by records of over 90 ICU patients with periodic measurements of their vital signs obtained from a bedside monitor. These measurements were taken over several hours of observation and the records contain metrics such as heart and breathing rates, ECG signals, blood pressure, SpO2, etc, and each patient is labeled with a clinical class. The clinical conditions available in this dataset are (for conciseness, ahead we refer to them by simply using the letter within brackets): \textbf{\textit{(A)}} Angina, \textbf{\textit{(B)}} Bleed, \textbf{\textit{(C)}} Brain Injury, \textbf{\textit{(D)}} Congestive Pulmonary Failure/Pulmonary Edema, \textbf{\textit{(E)}} Cardiac Arrest, \textbf{\textit{(F)}} Cardiogenic Shock, \textbf{\textit{(G)}} Post-OP Coronary Artery Bypass Graft (Post-OP CABG), \textbf{\textit{(H)}} Post-OP Valve, \textbf{\textit{(I)}} Renal Failure and \textbf{\textit{(J)}} Respiratory Failure.

\subsubsection{Data Preparation}
In our experiments, we intend to explore how we can augment and reshape the data collected from the healthy participants in such a way that it can represent a certain clinical class from the MIMIC-I dataset. Considering that the Zephyr BioModules used in the data collection are non-invasive, we can only collect signals such as heart rate and breathing rate with them, and not blood-related measurements. In this sense, after analysing the MIMIC-I dataset, we selected the maximum number of subjects that contain the Heart Rate (HR) and Breathing Rate (BR) together with other metrics that are consistent. This led us to a total of 50 different subjects that have recordings for the HR, BR, systolic ABP (Arterial Blood Pressure), mean ABP, diastolic ABP, Pulse and SpO2. 

Each subject was tracked over several hours, with most of them over 20 hours. To increase the sample size and make the time series suitable for our model, we sliced them into slices of 60 seconds (longer series would make the task more difficult to the model) and each was labeled with the respective clinical class of the subject (the slices of each subject were assigned with that subject's class). While other techniques such as Multiple Instance Learning could have been used for the label assignments and alleviate the model, we opted to label the sliced blocks individually and shuffle them for learning. This led us to over 70k sub-samples, each with a label out of the 10 mentioned above. The data was also normalised.

\begin{figure}
    \centering
    \includegraphics[width=\columnwidth]{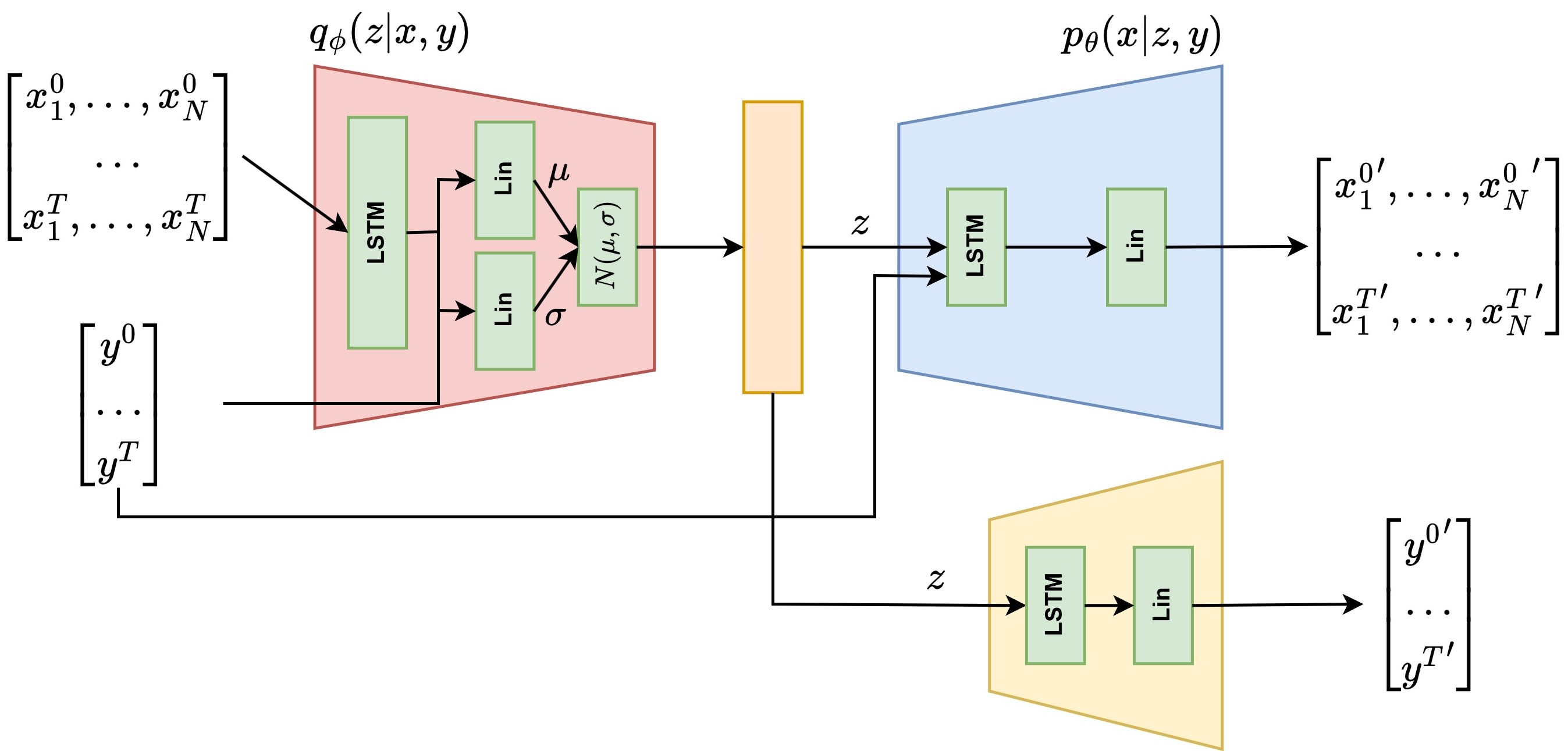}
    \caption{CVAE-inspired architecture of the proposed model in this paper to reshape time series samples from healthy individuals in such a way that mimics vital signs of unhealthy individuals.}
    \label{fig:model_arch}
\end{figure}
\section{Methodology}\label{sec:methods}
\subsection{Proposed Architecture - CVVitAE}
In this section, we propose Conditional Variational Vital Autoencoder (CVVitAE) (Fig. \ref{fig:model_arch}), a CVAE \cite{cvae_2015} model to reshape and augment the data collected from our healthy participants so that we can simulate as if they were being affected by a clinical condition present in the ICU database.
\begin{figure*}
    \centering
    \subfigure[Angina (A)]{\label{fig:env_a}\includegraphics[width=0.24\textwidth]{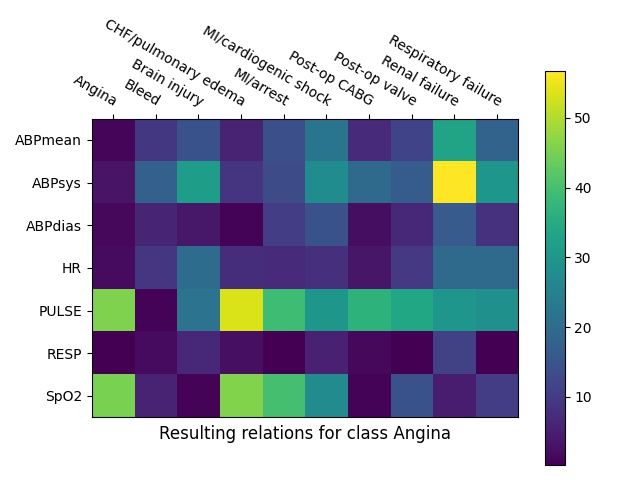}}
    \subfigure[Bleed (B)]{\label{fig:env_b}\includegraphics[width=0.24\textwidth]{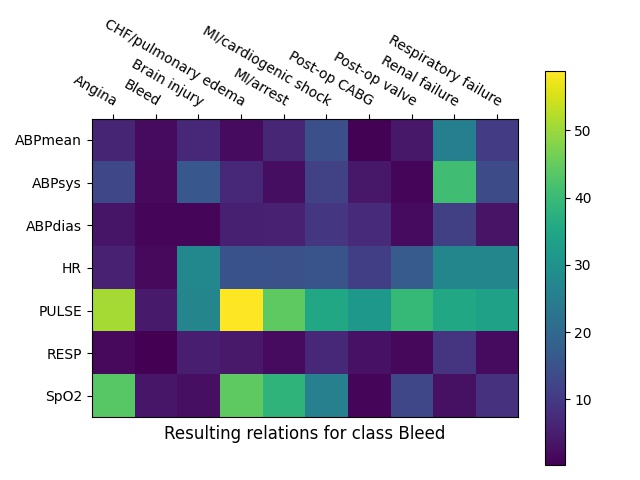}}
    \subfigure[MI Cardiogenic Shock (F)]{\label{fig:env_d}\includegraphics[width=0.24\textwidth]{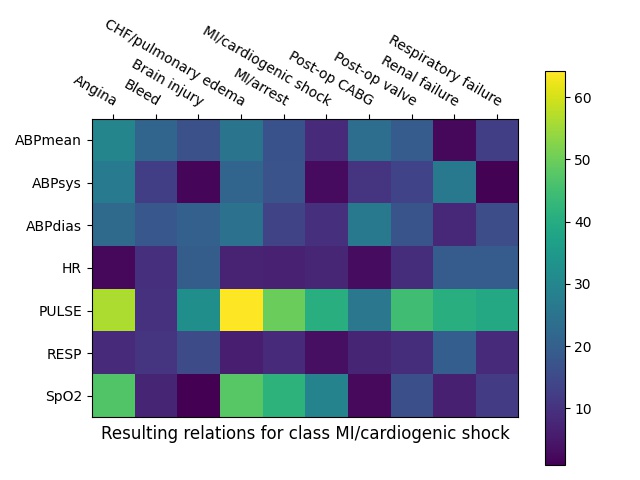}}
    \subfigure[Renal Failure (I)]{\label{fig:env_d}\includegraphics[width=0.24\textwidth]{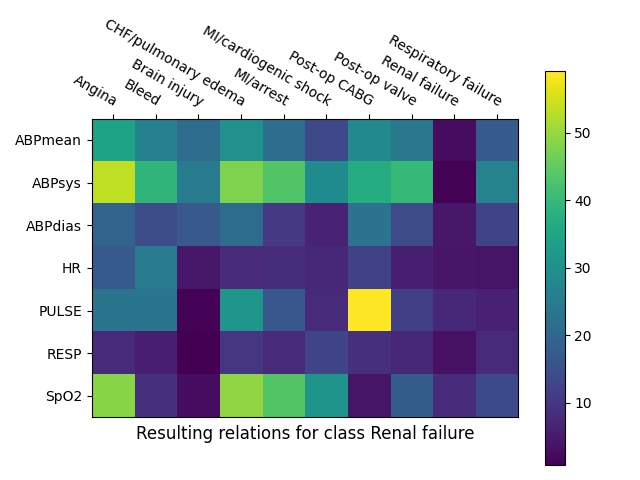}}
    \caption{Distances for the unhealthy samples generated by feeding our collected data from healthy individuals to the trained vanilla CVAE model. Darker colours represent smaller values. The letter within brackets represents the ID of the clinical class, as described in section \ref{sec:mimic-db}.}
    \label{fig:results_matrix_vanilla_cvae}
\end{figure*} 
\begin{figure*}
    \centering
    \subfigure[Angina (A)]{\label{fig:env_a}\includegraphics[width=0.24\textwidth]{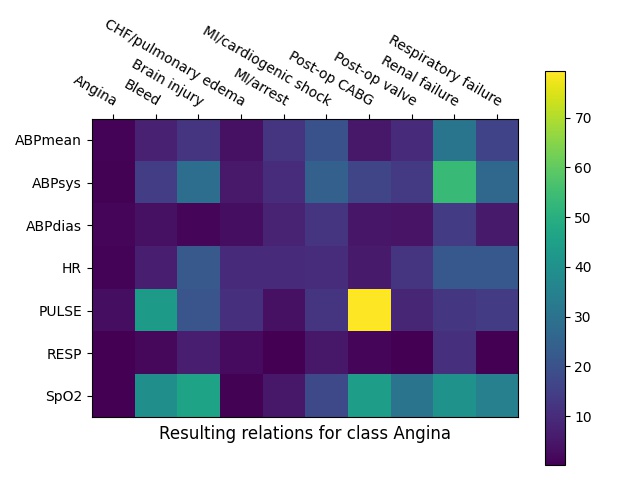}}
    \subfigure[Bleed (B)]{\label{fig:env_b}\includegraphics[width=0.24\textwidth]{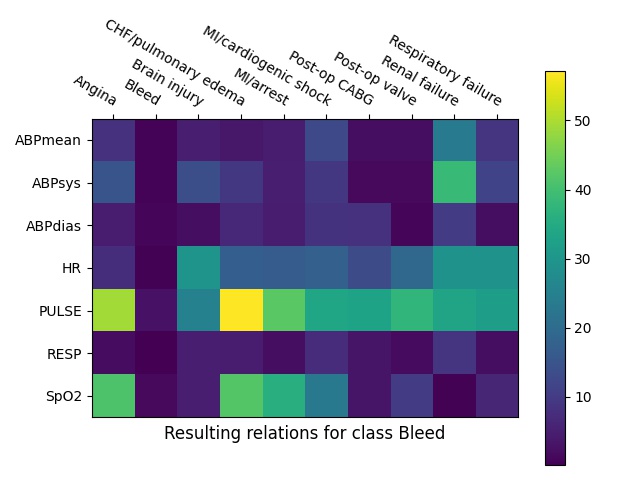}}
    \subfigure[MI Cardiogenic Shock (F)]{\label{fig:env_d}\includegraphics[width=0.24\textwidth]{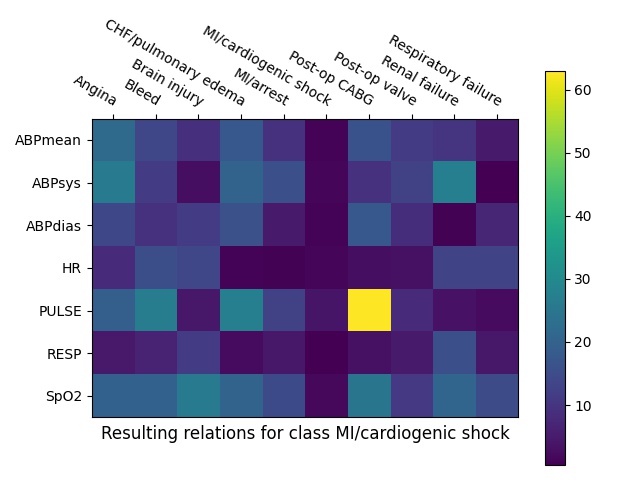}}
    \subfigure[Renal Failure (I)]{\label{fig:env_d}\includegraphics[width=0.24\textwidth]{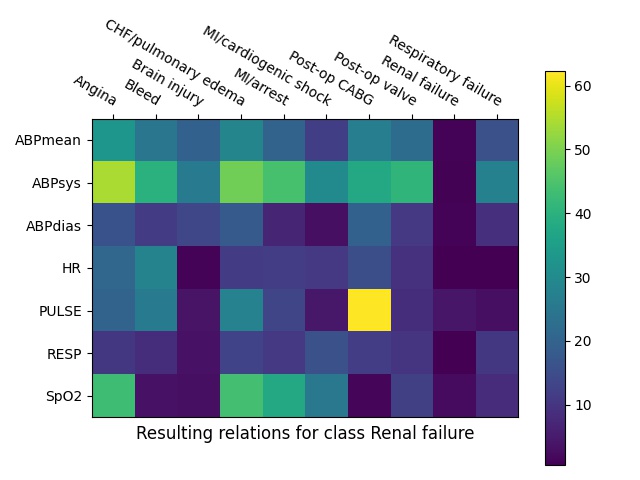}}
    \caption{Distances for the unhealthy samples generated by feeding our collected data from healthy individuals to the trained CVVitAE model. Darker colours represent smaller values. The letter within brackets represents the ID of the clinical class, as described in section \ref{sec:mimic-db}.}
    \label{fig:results_matrix_cvvitae}
\end{figure*}

After cleaning the MIMIC-I data as described in section \ref{sec:preliminaries}, we feed it to our model that is composed of an encoder and decoder networks, as in standard VAEs \cite{kingma2013auto}. However, the objective is to reshape data in a way that approximates a given clinical condition, meaning that the model needs to learn the relationships between the vital signs and the corresponding clinical labels. Hence, we follow the principles of CVAEs instead of normal VAEs. This means that the vital signs will be conditioned in the clinical labels before being fed to the model, i.e., $z = q_{\phi}(z|x,y)$, considering an encoder network $q_{\phi}$ that receives the inputs $x$ conditioned on the true labels $y$. The encoder will transform the inputs into the latent space $z$, from which new data points can be sampled and given to a decoder $p_{\theta}$ that will attempt to reconstruct the original input data. However, to control the values that are sampled from the latent space based on a given clinical class, the decoder can be represented following the probability distribution $p_{\theta}(x|z,y)$, where $y$ is a certain clinical label. Our network can be trained by minimising the ELBO loss as described in 
\begin{equation}\label{eq:elbo_loss}
    \mathcal{L_{\text{A}}}=\mathbb{E}_{q_{\phi(z|x,y)}}[\log p_\theta(x|z,y)]-D_\text{KL}[q_\phi(z|x,y)||p_{\theta}(z|x)].
\end{equation}
We approximate the reconstruction term with the MSE of the reconstructions from the network to the original inputs, which can be written as $\frac{1}{T}\sum_{t=1}^T(y'_t-y_t)^2$, and the KL-Divergence to a normal distribution $\mathcal{N}(0, I)$ for the regularisation term.

This describes the architecture of the CVAE applied to our problem with vital signs time series. However, simply using this model may not be enough to ensure that the latent space $z$ can accurately consider the clinical labels as a condition to the generated samples. To make sure that the latent space is aligned with the clinical labels, we include an additional component in our architecture that works as a regularizer unit for the latent space. We use an LSTM-based layer that receives the time series sampled from the latent space and will match them with the respective clinical label. In essence, this layer works as a latent space regularizer that predicts the labels from latent samples, resulting in a second objective of minimising the cross-entropy between the predicted labels by this classifier and the true labels. This loss can be represented as
\begin{equation}\label{eq:ce_loss}
\mathcal{L_{\text{B}}}=-\sum_{i = 1}^Cy_i \ log \ p(y_i),
\end{equation}
where $C$ is the number of clinical labels, and $p$ is the distribution for the predicted labels. This second loss function can then be integrated in our first loss described in Eq. (\ref{eq:elbo_loss}), resulting in the overall loss
\begin{equation}\label{eq:general_loss}
\mathcal{L}=\alpha \times \mathcal{L_{\text{A}}} + \mathcal{L_{\text{B}}},
\end{equation}
where $\alpha$ is a weighting factor that controls the weight given to the losses when training the model. All the components of the proposed architecture are depicted in Fig. \ref{fig:model_arch}.
%\begin{figure}
%    \centering
%    \includegraphics[width=0.6\columnwidth]{resources/out.png}
%    \caption{Train and test losses during training of the proposed CVVitAE model using the MIMIC-I dataset.}
%    \label{fig:loss}
%\end{figure}

\subsection{Evaluating the Proposed Model}
After training our model, the goal is to use the collected data and feed it to the model described in the previous section to be augmented. Normally, the trained decoder would be enough for the generation stage, but in this case, we are using real collected data and not just sampled noise. Hence, the data needs to go through the encoder in order to be consistent with the model. Since some of the features of the collected data are missing when compared to the MIMIC-I dataset, we fill the missing values with noise that is normalised in accordance with the values of the collected data and the dataset. Then this can be fed to the decoder, together with the clinical label that we intend to mimic by reshaping the original data.

Importantly, the vital signs of clinical labels of different classes can be close to each other since the variations in the vital signs over time can be very small. Hence, it might be non-trivial to evaluate the quality of the samples generated. For that purpose, we define a metric of proximity from the generated samples to the real values of the input features of each clinical class. The idea is to compare the proximity of the values generated for each feature to the same feature of the samples with the same clinical label in the original MIMIC-I dataset. To achieve this, we calculate the average values of each feature in the original set and build a map containing the distance from this average to the average of the values in the features of the augmented samples. This procedure is done for each clinical class that we augment data for, resulting in a value $m_{j,c}$ that illustrates the proximity between a generated sample from our model and the real samples of the MIMIC-I dataset for a given class $c$, and for each vital sign feature $j$. Formally, this can be written 
\begin{equation}
    m_{j,c} = \frac{\Big(\{x'_j\}_{t=1}^T - \overline{x}_j\Big)}{T},
\end{equation}
where $j$ corresponds to the index of each feature in the vital signs time series, $T$ to the number of timesteps in the series, and $c \in \{1,\dots,C\}$ to the index of a certain clinical label. 

\section{Results}\label{sec:results}
The main objective of this work is to create a generative model that aligns samples of data collected from healthy individuals with a target clinical condition. We evaluate our model (after training convergence) using samples from healthy individuals and attempt to reshape them in a way that mimics the vital signs of clinical conditions. This can be seen in the plots in Fig. \ref{fig:results_matrix_vanilla_cvae} and \ref{fig:results_matrix_cvvitae}, showing the distances of each feature for four generated classes to the features of each class in the real data. In Fig. \ref{fig:results_matrix_cvvitae}, we observe that, for each generated class with CVVitAE, the values of the features are always closer to the feature of that intended class in the initial dataset. This observation is consistent across all clinical classes that we have generated, meaning that our model is capable of augmenting samples of healthy individuals in a way that is aligned with an intended clinical condition. Instead, when using a vanilla CVAE the features are much more distant, as seen in Fig. \ref{fig:results_matrix_vanilla_cvae}. 

One of the key components of our architecture that potentiates the generation of aligned samples is the use of the latent space classifier that aligns the learned latent space with the intended clinical labels, as it can be seen in Table \ref{tab:tab1}, where the architecture that doesn’t use this classifier (Vanilla CVAE, similar to as used in \cite{cvae_paper_2023,vae_paper_2022}) generates samples that are more distant from the real data. The confusion matrix of the classifier layer from our architecture that matches the number of predicted with true labels can also be seen in Table \ref{tab:tab2}. We can see that the classification layer can accurately align the latent space of our CVVitAE with the intended labels. The use of LSTM layers in our architecture is also an important component for the success of our model, as it can be seen in Table \ref{tab:tab1}, where the samples generated from the same model that uses the classifier but with linear layers instead of LSTM layers in the encoder and decoder (Linear CVVitAE) generates samples that are more distant from the real ICU data.

Overall, our results demonstrate that the proposed model can successfully learn the dynamics of the used dataset and further process real data that we collected from healthy individuals and reshape it in a way that mimics as if they were suffering from some clinical condition. Our model also proves to be stronger when compared to the baselines.

\begin{table}[!t]
\caption{Distances to the clinical labels from our collected data samples augmented with the trained models.}
\begin{center}
\resizebox{!}{15pt}{
\begin{tabular}{|c|c|c|c|c|c|c|c|c|c|c|c|}
\hline
& \textbf{\textit{A}}& \textbf{\textit{B}}& \textbf{\textit{C}}& \textbf{\textit{D}}& \textbf{\textit{E}}& \textbf{\textit{F}}& \textbf{\textit{G}}& \textbf{\textit{H}}& \textbf{\textit{I}}& \textbf{\textit{J}} \\
\hline
Vanilla CVAE & 99.10 & 15.39 & 48.81 & 98.12 & 86.75 & 102.37 & 42.38 & 88.24 & 31.49 & 49.79 \\
\hline
Linear CVAE & 9.49 & 10.47 & 8.05 & 9.99 & 16.86 & 9.92 & 28.65 & 14.90 & 8.90 & 13.72 \\
\hline
CVVitAE & 6.89 & 7.34 & 13.61 & 9.98 & 7.21 & 11.99 & 9.46 & 18.99 & 11.31 & 12.09\\
\hline
\end{tabular}
}
\label{tab:tab1}
\end{center}
\end{table}

\begin{table}[!t]
\caption{Confusion matrix outputted by the classifier layer in our model at the end of training with the test set (15811 samples).}
\begin{center}
\resizebox{!}{50pt}{
\begin{tabular}{|c|c|c|c|c|c|c|c|c|c|c|}
\hline
\textbf{True}&\multicolumn{10}{|c|}{\textbf{Predicted Label}} \\
\cline{2-11} 
\textbf{Label} & \textbf{\textit{A}}& \textbf{\textit{B}}& \textbf{\textit{C}}& \textbf{\textit{D}}& \textbf{\textit{E}}& \textbf{\textit{F}}& \textbf{\textit{G}}& \textbf{\textit{H}}& \textbf{\textit{I}}& \textbf{\textit{J}} \\
\hline
\textbf{\textit{A}}& 1551 & 3 & 5 & 4 & 1 & 7 & 2 & 4 & 4 & 4  \\
\hline
\textbf{\textit{B}}& 4 & 2233 & 7 & 9 & 0 & 8 & 0 & 2 & 0 & 0  \\
\hline
\textbf{\textit{C}}& 12 & 7 & 788 & 3 & 2 & 3 & 0 & 0 & 1 & 0  \\
\hline
\textbf{\textit{D}}& 9 & 13 & 4 & 3241 & 0 & 0 & 5 & 3 & 0 & 16  \\
\hline
\textbf{\textit{E}}& 0 & 0 & 1 & 0 & 101 & 2 & 0 & 2 & 0 & 0  \\
\hline
\textbf{\textit{F}}& 1 & 4 & 3 & 0 & 1 & 1840 & 0 & 2 & 2 & 4  \\
\hline
\textbf{\textit{G}}& 6 & 0 & 1 & 8 & 0 & 0 & 503 & 3 & 1 & 8  \\
\hline
\textbf{\textit{H}}& 3 & 0 & 1 & 8 & 3 & 2 & 2 & 962 & 0 & 1  \\
\hline
\textbf{\textit{I}}& 6 & 0 & 1 & 0 & 1 & 6 & 2 & 5 & 505 & 9  \\
\hline
\textbf{\textit{J}}& 2 & 0 & 0 & 14 & 0 & 10 & 6 & 5 & 5 & 3804  \\
\hline
\end{tabular}
}
\label{tab:tab2}
\end{center}
\end{table}

\section{Conclusion and Further Work}\label{sec:conclusion}
It can be challenging to find representative data in applications such as healthcare in extreme environments as in battlefields, natural disasters or sports, which are needed to train machine learning models. In this paper, we proposed a CVAE architecture that can process ICU data and generate new time series of sensor signals that relate to a targeted clinical condition. In this sense, we can use the vital signs data collected from healthy individuals and augment these with patterns associated with targeted clinical conditions. Leveraging generative AI to help process and augment this type of signals can be crucial to ensure feature consistency across despaired data sources. By learning latent signal representations it is possible to mitigate these challenges.

Overall, the results showed that the proposed generative architecture is capable of aligning samples with the intended clinical condition with 98\% accuracy. Furthermore, the proposed method proved to be successful in aligning vital signs from healthy individuals with the targeted clinical conditions, as evaluated by our metric of distance to the real samples. In the future, we intend to include more modalities in the proposed method such as visual signals and localisation, helping to scale it to more complex scenarios. 

%\section*{Acknowledgment}
%This work was funded by the Engineering and Physical Sciences Research Council, grant number EP/X028631/1: ATRACT: A Trustworthy Robotic Autonomous system to support Casualty Triage.

\bibliography{references}
\bibliographystyle{IEEEtran}

\end{document}